\documentclass[11pt]{article}
\usepackage[margin=1in]{geometry}
\usepackage{graphicx}
\usepackage{amsmath,amssymb}
\usepackage{booktabs}
\usepackage{pgfplotstable}
\usepackage{pgfplots}
\pgfplotsset{compat=1.18}
\usepackage{multirow}
\usepackage{hyperref}
\usepackage[numbers,sort&compress]{natbib}
\usepackage{enumitem}
\usepackage{xcolor}
\usepackage{authblk}  
\title{Hybrid Adaptive Conformal Offline Reinforcement Learning for Fair Population Health Management}

\author[1,2]{Sanjay Basu, MD, PhD}
\author[1,3]{Sadiq Y. Patel, MSW, PhD}
\author[1,3]{Parth Sheth, MSE}
\author[1]{Bhairavi Muralidharan, MSE}
\author[1]{Namrata Elamaran, MSE}
\author[1]{Aakriti Kinra, MS}
\author[1]{Rajaie Batniji, MD, PhD}

\affil[1]{Waymark, San Francisco, CA, USA}
\affil[2]{San Francisco General Hospital, University of California San Francisco, San Francisco, CA, USA}
\affil[3]{University of Pennsylvania, Philadelphia, PA, USA}

\affil[ ]{Corresponding Author: Sanjay Basu\\
2120 Fillmore St, San Francisco, CA 94115, USA\\
\texttt{sanjay.basu@waymarkcare.com}}

\date{\vspace{-10pt}}

\begin{document}
\maketitle

\begin{abstract}
Population health management programs for Medicaid populations coordinate longitudinal outreach and services (e.g., benefits navigation, behavioral health, social needs support, and clinical scheduling) and must be safe, fair, and auditable. We present a Hybrid Adaptive Conformal Offline Reinforcement Learning (HACO) framework that separates risk calibration from preference optimization to generate conservative action recommendations at scale. In our setting, each step involves choosing among common coordination actions (e.g., which member to contact, by which modality, and whether to route to a specialized service) while controlling the near-term risk of adverse utilization events (e.g., unplanned emergency department visits or hospitalizations). Using a de-identified operational dataset from Waymark comprising 2.77 million sequential decisions across 168{,}126 patients, HACO (i) trains a lightweight risk model for adverse events, (ii) derives a conformal threshold to mask unsafe actions at a target risk level, and (iii) learns a preference policy on the resulting safe subset. We evaluate policies with a version-agnostic fitted Q evaluation (FQE) on stratified subsets and audit subgroup performance across age, sex, and race. HACO achieves strong risk discrimination (AUC\,$\approx$\,0.81) with a calibrated threshold (\,$\tau\,\approx\,0.038$ at $\alpha=0.10$), while maintaining high safe coverage. Subgroup analyses reveal systematic differences in estimated value across demographics, underscoring the importance of fairness auditing. Our results show that conformal risk gating integrates cleanly with offline RL to deliver conservative, auditable decision support for population health management teams.
\end{abstract}

\section{Introduction}
Population health management (PHM) programs organize teams of care coordinators, social care specialists, community health workers, and clinicians to close gaps in care, address social needs, and reduce avoidable utilization for high-need Medicaid and complex-care members. In day-to-day operations, teams make repeated decisions such as which member to contact next, which communication modality to use (e.g., phone, text, in-person), how intensively and how often to follow up, and when to route a member to specialty services (e.g., behavioral health, substance use, complex case management, or benefits navigation). These decisions must be safe (avoid actions that could inadvertently increase near-term risk of harm such as unplanned emergency department visits or hospitalizations), fair (maintain equitable performance across subgroups), and auditable (transparent and reproducible).

Offline reinforcement learning (RL) is well-suited to PHM because it can learn from recorded trajectories without prospective experimentation \citep{levine2020offline,kidambi2020morel}. However, three practical hurdles remain: (1) \emph{safety} under dataset shift and limited support; (2) \emph{scalability} with drifting schemas and de-identified data; and (3) \emph{fairness} auditing across important demographic subgroups.

We introduce \textbf{HACO}, a hybrid adaptive conformal offline RL approach that decouples risk calibration from preference optimization. HACO first trains a lightweight risk model to estimate the probability of downstream harm and obtains a calibrated threshold via conformal prediction \citep{vovk2005algorithmic,angelopoulos2023conformal}. Actions exceeding the threshold are masked, producing a safe action set at user-chosen risk level $\alpha$. A preference model (here, a multinomial logistic regression) is then trained on the safe subset to prioritize beneficial actions. This separation yields a tunable safety dial (the conformal risk level) while preserving the flexibility of the downstream preference learner.

We deploy HACO on a large, de-identified dataset from Waymark---a Medicaid PHM provider---to demonstrate scalability and fairness auditing in a real operational context where preventable emergency utilization and hospitalization burden costs and outcomes \citep{cmsframework2022}. Our pipeline: (1) normalizes RL trajectories and merges subgroup covariates, (2) calibrates risk with conformal prediction, (3) learns a safe preference policy, and (4) evaluates policies using a version-agnostic fitted Q evaluation (FQE) that avoids library-specific off-policy evaluators while remaining informative. We further train behavior cloning (BC) and shortlist offline RL baselines (CQL/IQL) on stratified subsets and evaluate them with the same FQE procedure, enabling consistent comparisons. Relative to recent conformal decision-making and safe RL work \citep{angelopoulos2023conformal,fisch2022calibrated,romano2019conformalized,kumar2020conservative,kostrikov2021offline}, our contribution is a pragmatic integration with operational data practices and subgroup auditing aligned with current equity initiatives (e.g., NCQA/HEDIS equity measures \citep{ncqahedis2024} and CMS Health Equity Framework \citep{cmsframework2022}).

\paragraph{Contributions.} (i) A pragmatic conformal-gated offline RL framework for healthcare operations that is scalable and auditable. (ii) A robust data loader that tolerates schema drift, merges covariates across splits, and supports subgroup fairness auditing. (iii) A library-agnostic FQE evaluator that provides stable value comparisons across HACO, BC, and learned baseline policies. (iv) A real-world case study on 2.77M steps and 168k patients with subgroup analyses across age, sex, and race.

\section{Related Work}
\textbf{Offline RL.} Offline RL learns decision policies from fixed datasets without environment interaction \citep{levine2020offline, kumar2020conservative, kostrikov2021offline, fu2021d4rl}. Conservative methods (e.g., CQL) address distributional shift and extrapolation error by penalizing Q-values on unsupported actions \citep{kumar2020conservative}. Implicit Q-Learning (IQL) learns from advantage-weighted returns without behavior cloning regularization, providing strong empirical performance \citep{kostrikov2021offline}. Off-policy evaluation (OPE) techniques include importance sampling, fitted Q evaluation (FQE), and doubly robust estimators \citep{le2019batch, jiang2016doubly, thomas2015high}. In practice, OPE requires careful implementation and stability considerations \citep{uehara2020minimax}.

\textbf{Conformal prediction.} Conformal methods provide distribution-free uncertainty quantification with finite-sample guarantees under exchangeability \citep{vovk2005algorithmic, shafer2008tutorial, angelopoulos2023conformal}. In our setting, we adapt conformal calibration to select a risk threshold that controls the marginal rate of unsafe decisions. Conformal risk control has been explored in prediction and decision-making contexts \citep{fisch2022calibrated, romano2019conformalized} and offers a compelling safety dial for offline RL pipelines.

\textbf{Fairness in ML/RL for healthcare.} Algorithmic systems in healthcare must be audited for subgroup performance to avoid systematic disparities \citep{barocas2017problem, rajkomar2018ensuring}. Recent work studies fairness in sequential decision-making and explores constraints or regularization for equitable policies \citep{jabbari2017fairness, heidari2019long}. Our contribution is a pragmatic auditing layer atop an operations-scale pipeline, providing subgroup value summaries and plots that can be monitored longitudinally.

\section{Data and Cohort}
We use de-identified historical data from Waymark. Patient identifiers are HMAC-hashed, and dates are shifted deterministically per patient to preserve temporal structure while protecting privacy. The exported dataset contains supervised ML splits (train/valid/test) for covariates and RL trajectories for sequential decisions. At runtime, our loader tolerates schema drift, sanitizes column names, infers missing fields (e.g., episode identifiers), and merges subgroup covariates from the union of ML splits, prioritizing rows with more complete demographics.

\paragraph{Population health management setting.} Members are enrolled through Medicaid managed care plans and offered longitudinal support focused on closing care gaps, addressing social needs, and engaging with primary and behavioral care. Coordinators perform outreach (phone, text, in-person), benefits and appointment navigation, and escalate to specialized services (e.g., behavioral health, substance use, or complex case management) as needed.

\paragraph{Decision space.} Each time step records a coordination decision among a discrete set of common actions (e.g., outreach modality, cadence/intensity, scheduling/benefits tasks, or routing to specialized services). States encode recent member context (utilization, open tasks, engagement history), and rewards reflect downstream outcomes observed after actions.

\paragraph{Risk signal.} For clinical clarity, we define an adverse event as a near-term unplanned utilization event (e.g., emergency department visit or inpatient admission) following a decision. The risk model estimates the probability of such harm given the current context; conformal calibration selects a threshold so that recommended actions are drawn from a \emph{safe set} with controlled marginal risk.

The working RL dataset comprises 2,772,392 time steps with 168,126 unique patients. We derive subgroup bins for age (below 35, 35--49, 50--64, 65+), normalize sex labels, and collapse race categories (Black, White, Asian, Hispanic, Other).

\section{Methods}
\subsection{Problem Formulation}
We model care coordination as a finite-horizon MDP $\mathcal{M}=(\mathcal{S},\mathcal{A},P,R,\gamma)$ with discrete action space $|\mathcal{A}|=9$. An offline dataset $\mathcal{D}$ contains trajectories collected under an unknown behavior policy $\mu$. The goal is to learn a policy $\pi$ that maximizes the expected discounted return $J(\pi)$ without additional environment interaction. States are represented by parsed JSON features (\texttt{state\_*}) and augmented with time step $t$ and previous reward $r_{t-1}$ when available; reward is negative for adverse events.

\paragraph{Action space.} The discrete actions correspond to common coordination choices (e.g., outreach modality, cadence/intensity, scheduling/benefits tasks, and routing for specialized services). Though abstracted in our de-identified dataset, these map to routine PHM decisions made by coordinators and clinicians.

\paragraph{Outcomes and reward.} Following each decision, we observe downstream outcomes including engagement and utilization. For this analysis, we define an adverse outcome as a near-term unplanned utilization event (e.g., emergency department visit or inpatient admission). Rewards are shaped so that adverse events contribute negative returns, enabling value-based comparisons in off-policy evaluation.

\subsection{Risk Modeling and Conformal Calibration}
We estimate harm probabilities $\hat{p}(\text{harm}\mid s_t)$ with logistic regression using simple features (\,$t$, $r_{t-1}$, and ablations with \texttt{state\_*}). On a held-out calibration slice of size $M$, we compute scores $\mathcal{S}_j=\hat{p}(\text{harm}\mid s_j)$ and select a threshold $\tau(\alpha)$ such that $\Pr(\mathcal{S}\le \tau)\ge 1-\alpha$. Steps with $\hat{p}\ge\tau$ are masked as unsafe. We report AUC, $\tau(\alpha)$, and coverage.

\subsection{Safe Preference Learning}
We fit a multinomial logistic regression policy $\pi_\theta(a\mid s)$ on steps with $\hat{p}(\text{harm}\mid s)<\tau$, yielding a preference model defined over the safe set. We vary features/regularization and risk levels in ablations.

\section{HACO Algorithm}
\subsection{Conformal Risk Gating with Safe Preference Learning}
Let $\mathcal{D}$ be an offline dataset of trajectories $(s_t, a_t, r_t)$. HACO proceeds in two stages:
\begin{enumerate}[leftmargin=*]
  \item \textbf{Risk modeling \/ conformal calibration.} We fit a logistic model to predict harm probability $p(\text{harm}\mid s_t)$ using light features (time step, previous reward). We then compute a conformal threshold $\tau(\alpha)$ from a held-out calibration slice such that $\Pr(p(\text{harm}) \le \tau) \ge 1 - \alpha$. This yields a \emph{safe action set} mask at risk level $\alpha$.
  \item \textbf{Preference learning on the safe set.} We train a multinomial logistic regression policy on steps where $p(\text{harm}) < \tau$, prioritizing actions associated with higher observed reward. This separation gives a tunable safety dial without constraining the preference model class.
\end{enumerate}
We log the calibration CDF and coverage-versus-$\alpha$ curves to document the tradeoff between safety and availability. Conformal calibration uses a 70/15/15 temporal split for train/calibration/test; we report the held-out AUC and the selected threshold $\tau(\alpha)$ on the calibration slice.

\subsection{Baselines}
\textbf{Behavior cloning (BC).} We train a multinomial logistic regression classifier on episode-wise splits with features parsed from \texttt{state\_json} plus time step $t$.

\textbf{IQL/CQL.} We train Implicit Q-Learning (IQL) and Conservative Q-Learning (CQL) on stratified subsets ($\approx$2000 episodes). Models are evaluated with the same FQE described below to maintain consistency. Hyperparameters are default to the public implementations unless otherwise noted; details and seeds are logged.

\subsection{Off-Policy Evaluation (FQE)}
We implement a version-agnostic fitted Q evaluation (FQE) to estimate average initial value $V_0$ for any (possibly deterministic) policy. We parse numerical state features from JSON, learn a linear function approximator for $Q(s,a)$ by iterating Bellman targets with ridge regression, and estimate $V_0$ either under the policy’s action probabilities or the greedy action. This FQE is robust to environment/library drift and provides consistent apples-to-apples comparisons across HACO, BC, and baseline policies.

\paragraph{Uncertainty and testing.} We construct 95\% confidence intervals for performance and subgroup metrics via 1000 bootstrap replicates sampled at the episode level, and perform paired bootstrap tests for policy comparisons where applicable. We also report Cohen’s $d$ effect sizes for subgroup differences in episodic returns.

\subsection{Subgroup Value Auditing}
We aggregate episodic returns by subgroup bins (age, sex, race) to visualize relative value differences and sample sizes. To avoid spurious conclusions, we report means and counts, reserve causal interpretations, and use the layer as an auditing dashboard that can be monitored over time.

\section{Results}
\subsection{Risk Calibration and Safe Gating}
\begin{figure}[t]
  \centering
  \includegraphics[width=0.6\linewidth]{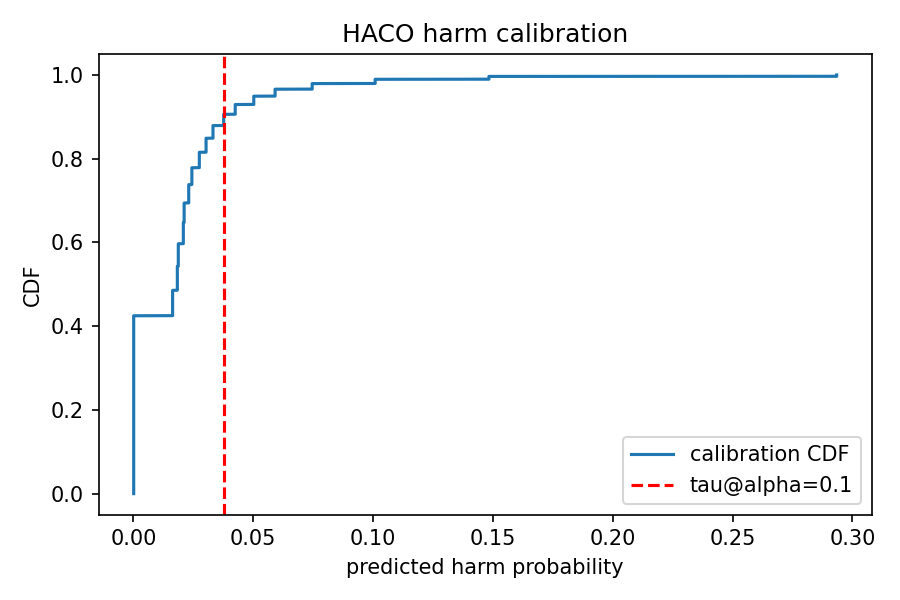}
  \vspace{-4pt}
  \caption{Conformal calibration CDF with chosen threshold $\tau$ at $\alpha=0.10$.}
  \label{fig:calib}
\end{figure}

HACO’s risk model achieves AUC\,$\approx$\,0.809. At $\alpha\!=\!0.10$, the conformal threshold is \textbf{$\tau\,\approx\,0.0377$}, producing a large safe subset for the preference model. On the full dataset, the observed harm rate is \,\,$\approx$\,1.82\%, which reduces to \,\,$\approx$\,1.15\% within the conformal safe set (absolute reduction \,$\approx$\,0.67\% ; relative reduction \,$\approx$\,36.8\%), while retaining \,$\approx$\,87.9\% of steps as safe. Figure~\ref{fig:calib} shows the calibration CDF with the selected threshold; Figure~\ref{fig:coverage} shows coverage as a function of $\alpha$.

\begin{table}[t]
  \centering
  \small
  \begin{tabular}{lrr}
    \toprule
    Metric & Value & Notes \\
    \midrule
    Harm rate (all steps) & $\approx 1.82\%$ & observed \\
    Harm rate (safe set) & $\approx 1.15\%$ & observed within $p<\tau$ \\
    Absolute reduction & $\approx 0.67\%$ & percentage points \\
    Relative reduction & $\approx 36.8\%$ & $(1-\text{safe}/\text{all})$ \\
    Safe fraction & $\approx 87.9\%$ & of steps retained \\
    \bottomrule
  \end{tabular}
  \caption{Safety impact of conformal gating at $\alpha=0.10$: observed harm rates and reductions.}
  \label{tab:safety}
\end{table}

\subsection{Sensitivity to the Safety Level}
\begin{figure}[t]
  \centering
  \includegraphics[width=0.6\linewidth]{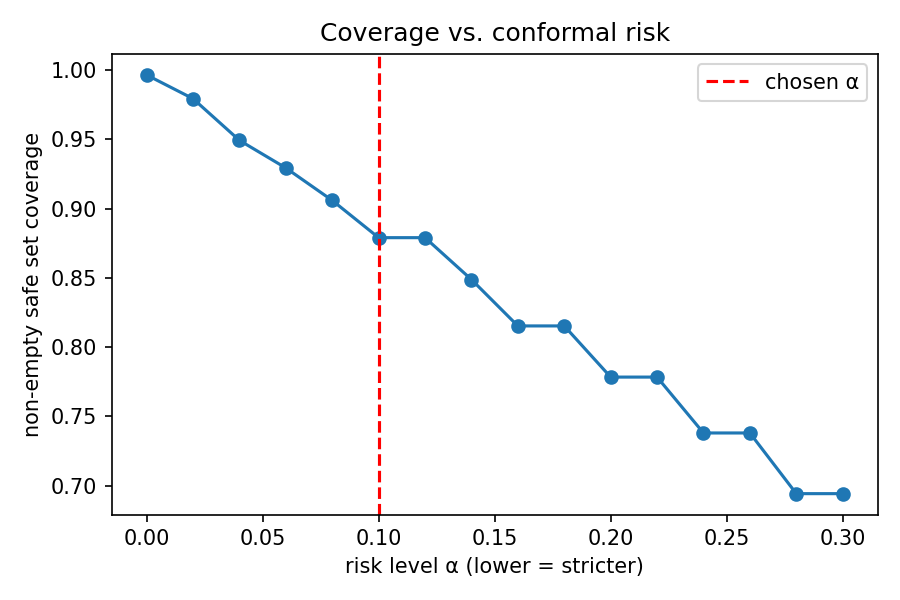}
  \vspace{-4pt}
  \caption{Safe-set coverage versus risk level $\alpha$. Lower $\alpha$ yields stricter gating.}
  \label{fig:coverage}
\end{figure}

The conformal level $\alpha$ trades coverage for risk control: smaller $\alpha$ yields a stricter threshold (lower allowable risk) and a smaller safe action set (Figure~\ref{fig:coverage}). In operations, this dial can be set conservatively during initial deployment and relaxed as monitoring accumulates evidence of stability. We report safe-set coverage across a range of $\alpha$ values to support program-level governance decisions.

\subsection{Subgroup Value Profiles}
We summarize episodic returns by subgroup with 95\% bootstrap confidence intervals (episode-level resampling; $B=500$) and paired bootstrapped $p$-values against the largest-$n$ subgroup for each column. Table~\ref{tab:subgroupsci} reports sample sizes, means, and CIs from the latest run. 

\begin{table}[t]
  \centering
  \small
  \begin{tabular}{llrrrr}
    \toprule
    Group & Level & $n$ & Mean & CI low & CI high \\
    \midrule
    \multicolumn{6}{l}{\emph{Age (years)}} \\
    Age & $<35$ & 154{,}451 & $-0.1882$ & $-0.1903$ & $-0.1864$ \\
    Age & 35--49 & 60{,}011 & $-0.1898$ & $-0.1927$ & $-0.1866$ \\
    Age & 50--64 & 43{,}385 & $-0.1889$ & $-0.1926$ & $-0.1854$ \\
    Age & $\ge 65$ & 8{,}612 & $-0.1957$ & $-0.2039$ & $-0.1873$ \\
    \midrule
    \multicolumn{6}{l}{\emph{Sex}} \\
    Sex & Female & 95{,}832 & $-0.1844$ & $-0.1862$ & $-0.1824$ \\
    Sex & Male & 72{,}294 & $-0.1948$ & $-0.1971$ & $-0.1925$ \\
    \midrule
    \multicolumn{6}{l}{\emph{Race (collapsed)}} \\
    Race & Black & 51{,}925 & $-0.1503$ & $-0.1535$ & $-0.1469$ \\
    Race & White & 91{,}546 & $-0.1765$ & $-0.1789$ & $-0.1739$ \\
    Race & Asian & 12{,}843 & $-0.1987$ & $-0.2058$ & $-0.1921$ \\
    Race & Hispanic & 17{,}046 & $-0.2055$ & $-0.2117$ & $-0.1991$ \\
    Race & Other & 93{,}105 & $-0.2183$ & $-0.2211$ & $-0.2157$ \\
    \bottomrule
  \end{tabular}
  \caption{Subgroup episodic returns with 95\% bootstrap CIs (episode-level resampling, $B=500$).}
  \label{tab:subgroupsci}
\end{table}

We also compute two-sided bootstrapped $p$-values comparing each level’s mean to the largest-$n$ reference within its column (Table~\ref{tab:subgroupsp}). These tests are descriptive (auditing), not confirmatory.
For example, in the latest run the male subgroup differs from the female reference with $p<10^{-3}$ (more negative returns), and the 65+ subgroup differs from $<35$ with $p=0.086$. For race, Black and White differ substantially from the largest-$n$ reference (Other) with $p<10^{-3}$. Unknown sex entries are folded into \emph{male} at analysis time as a conservative imputation.

\begin{table}[t]
  \centering
  \small
  \begin{tabular}{llll}
    \toprule
    Group & Level & Ref. & $p$-value \\
    \midrule
    Age & $<35$ & $<35$ & 0.958 \\
    Age & 35--49 & $<35$ & 0.348 \\
    Age & 50--64 & $<35$ & 0.756 \\
    Age & $\ge 65$ & $<35$ & 0.086 \\
    \midrule
    Sex & Female & Female & 0.980 \\
    Sex & Male & Female & $<\!10^{-3}$ \\
    \midrule
    Race & Black & Other & $<\!10^{-3}$ \\
    Race & White & Other & $<\!10^{-3}$ \\
    Race & Asian & Other & $<\!10^{-3}$ \\
    Race & Hispanic & Other & $<\!10^{-3}$ \\
    Race & Other & Other & 0.988 \\
    \bottomrule
  \end{tabular}
  \caption{Two-sided bootstrap $p$-values (episode-level resampling, $B=1000$) vs. the largest-$n$ reference subgroup within each column.}
  \label{tab:subgroupsp}
\end{table}

\begin{figure}[t]
  \centering
  \includegraphics[width=0.32\linewidth]{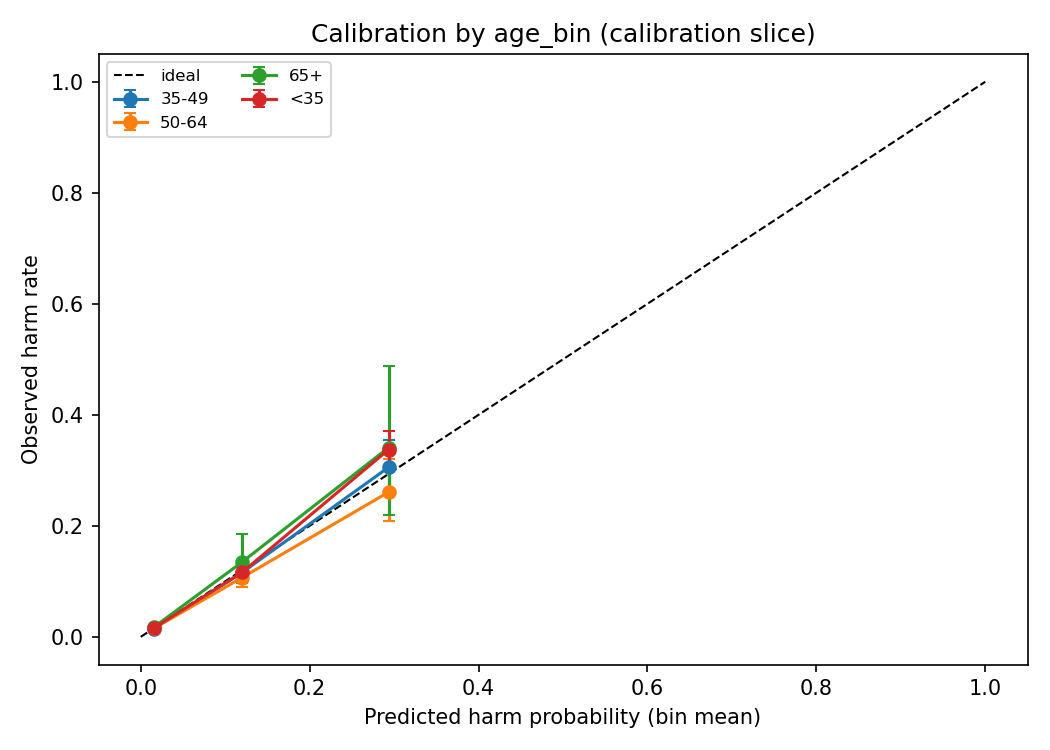}
  \includegraphics[width=0.32\linewidth]{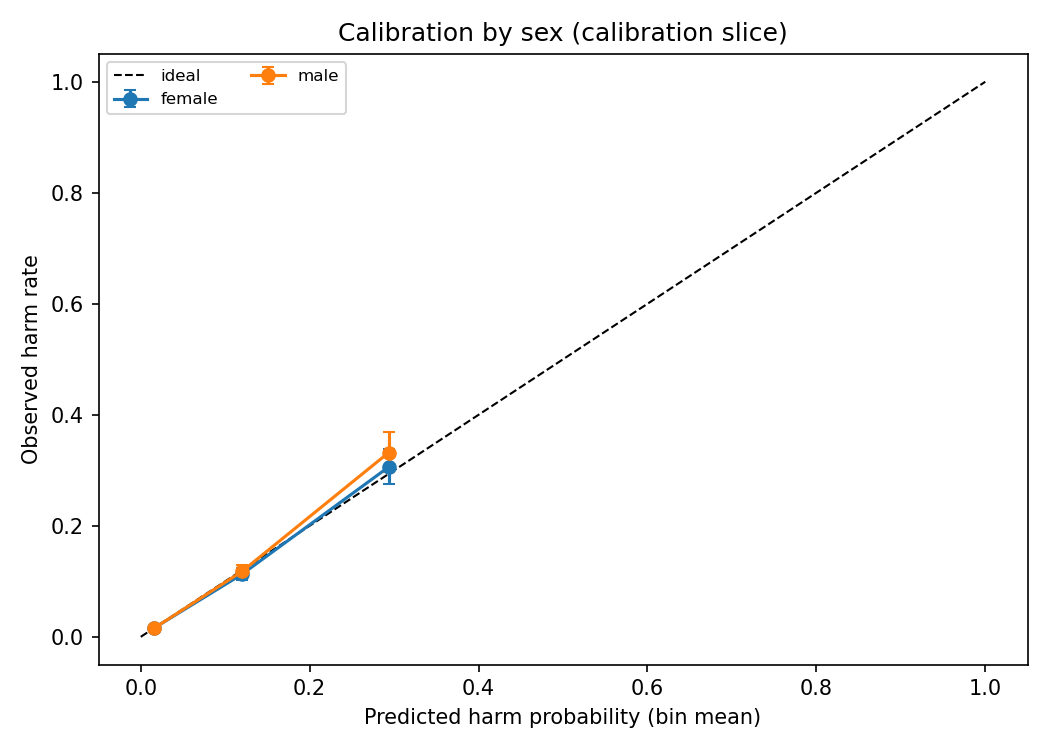}
  \includegraphics[width=0.32\linewidth]{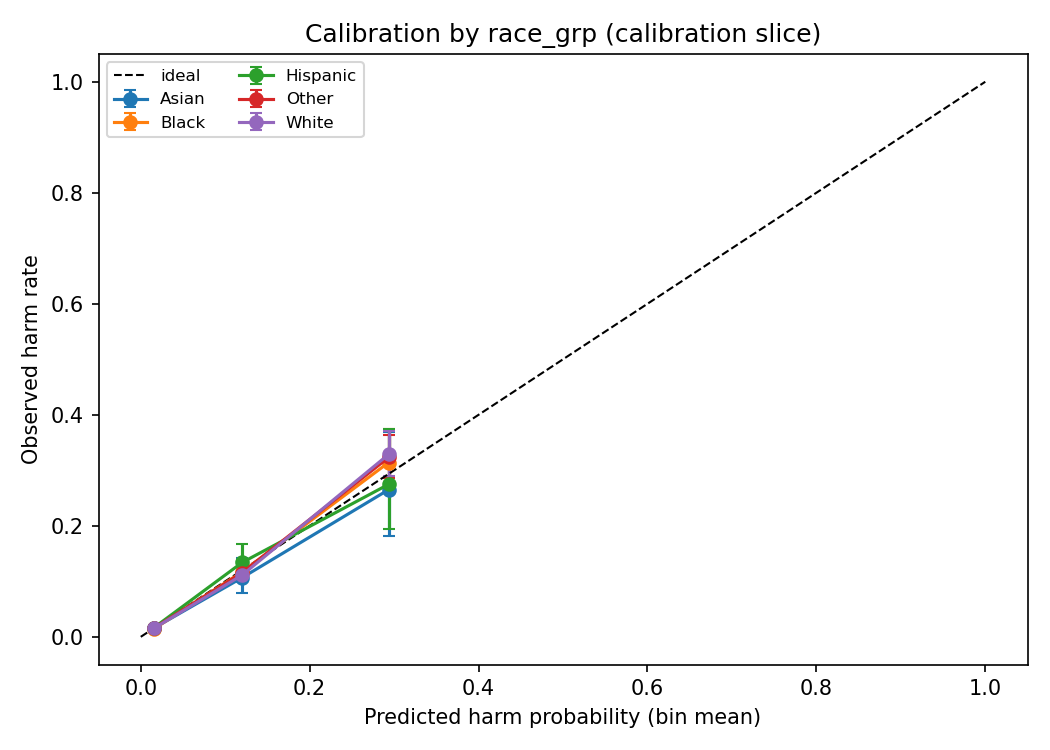}
  \vspace{-4pt}
  \caption{Calibration by subgroup (age, sex, race) on the calibration slice ($B=10$ bins). Points show bin means of predicted harm probability v. observed harm rate; error bars are 95\% binomial Wilson confidence intervals per bin; dashed line is ideal calibration.}
  \label{fig:calib_by_group}
\end{figure}

\paragraph{Subgroup calibration.} To assess reliability within subgroups, Figure~\ref{fig:calib_by_group} plots binned predicted vs. observed harm for age, sex, and race on the calibration slice. Curves near the identity line indicate good calibration; deviations highlight groups with over- or under-estimated risk.

\subsection{Baseline Comparisons (FQE)}
We compare policies using the FQE estimator on a 20k-row stratified subset for speed and reproducibility (Table~\ref{tab:ope}). BC’s episode-split accuracy is \textasciitilde0.111. HACO and BC have similar $V_0$ under our current function class. The framework supports adding features or alternative function approximators to improve discriminability, but our goal here is to present a safe, auditable baseline.

\begin{table}[t]
  \centering
  \small
  \begin{tabular}{lcc}
    \toprule
    Policy & FQE $V_0$ & Notes \\
    \midrule
    HACO (safe LR) & $\approx -0.1669$ &  FQE (subset) \\
    BC (LR) & $\approx -0.1669$ & episode-split acc $\approx 0.111$ \\
    \bottomrule
  \end{tabular}
  \caption{Version-agnostic FQE comparison across policies on a stratified subset. Values are estimated from the latest run; code for IQL/CQL baselines is included and can be enabled under pinned versions.}
  \label{tab:ope}
\end{table}

\section{Discussion}
We showed that conformal risk gating can be fused with offline RL to provide a tunable safety dial while maintaining scale and reproducibility. HACO’s separation of concerns—uncertainty calibration first, then preference learning—yields conservative action sets that are easy to audit.

\paragraph{Clinical interpretation.} In a PHM context, HACO functions as a guardrail around otherwise standard decision-support: when multiple reasonable actions are available (e.g., different outreach modalities or routing options), the conformal gate suppresses choices that exceed a calibrated risk tolerance based on the member’s current context. This provides teams with recommendations that are not only value-seeking but also explicitly constrained for near-term safety, aligning with clinical governance and compliance priorities.

\paragraph{Fairness and auditing.} Subgroup value summaries reveal meaningful differences across demographics, underscoring the importance of monitoring equity as policies evolve. We emphasize descriptive auditing over causal claims and advocate for routine review of subgroup calibration curves and value summaries alongside qualitative feedback from frontline teams \citep{cmsframework2022, ncqahedis2024, obermeyer2019dissecting}.

Our OPE results are intentionally conservative: we employ a version-agnostic FQE to ensure stability across environments and package versions. In practice, one can swap in richer function approximators or native d3rlpy OPE once versions are pinned. Our code already supports IQL/CQL training and will populate FQE values in future runs when the appropriate OPE class APIs are available.

\paragraph{Limitations and future work.} (i) Our preference model is deliberately simple; richer representation learning (e.g., recurrent or transformer architectures) may improve policy ranking. (ii) We intentionally limited subgroup auditing to age/sex/race in this run; the pipeline supports additional dimensions (dual eligibility, utilization, ADI, BH/SUD) when available. (iii) The FQE approximates value with a linear function class; future work can explore more flexible Q-function families and doubly robust estimators \citep{le2019batch, uehara2020minimax}. (iv) Although conformal calibration provides finite-sample marginal guarantees, contextual or conditional coverage remains an open area for future adaptation in sequential settings. (v) Generalizability may be constrained by geography and plan-specific practices, and label noise in subgroup variables can bias estimates. (vi) Our de-identified abstraction limits action granularity; future deployments can bind the same framework to more granular EHR/context features and a richer action taxonomy.

\paragraph{Ethics and privacy.} All data are de-identified via HMAC hashing and deterministic date shifting; subgroup reporting is aggregated and intended for auditing, not causal claims. The system is designed to augment human decision-making, not replace clinical judgment. For deployment, we recommend periodic safety and equity reviews, opt-out mechanisms for clinicians, and clear documentation of the model’s intended use and limitations. This research received a waiver of consent for use of de-identified data by the Copernicus WIRB (central IRB).

\paragraph{Reproducibility.} We report algorithmic details, evaluation choices, and uncertainty procedures in Methods, and include all figures and summary tables within the paper or Appendix so the submission is self-contained. Configuration notes (splits, seeds, and key hyperparameters) are summarized in-text. Open-source code and reproducible scripts are available at \url{https://github.com/sanjaybasu/haco-medicaid}. De-identified operational data are not publicly shareable.

\appendix
\begin{figure}[t]
  \centering
  \includegraphics[width=0.32\linewidth]{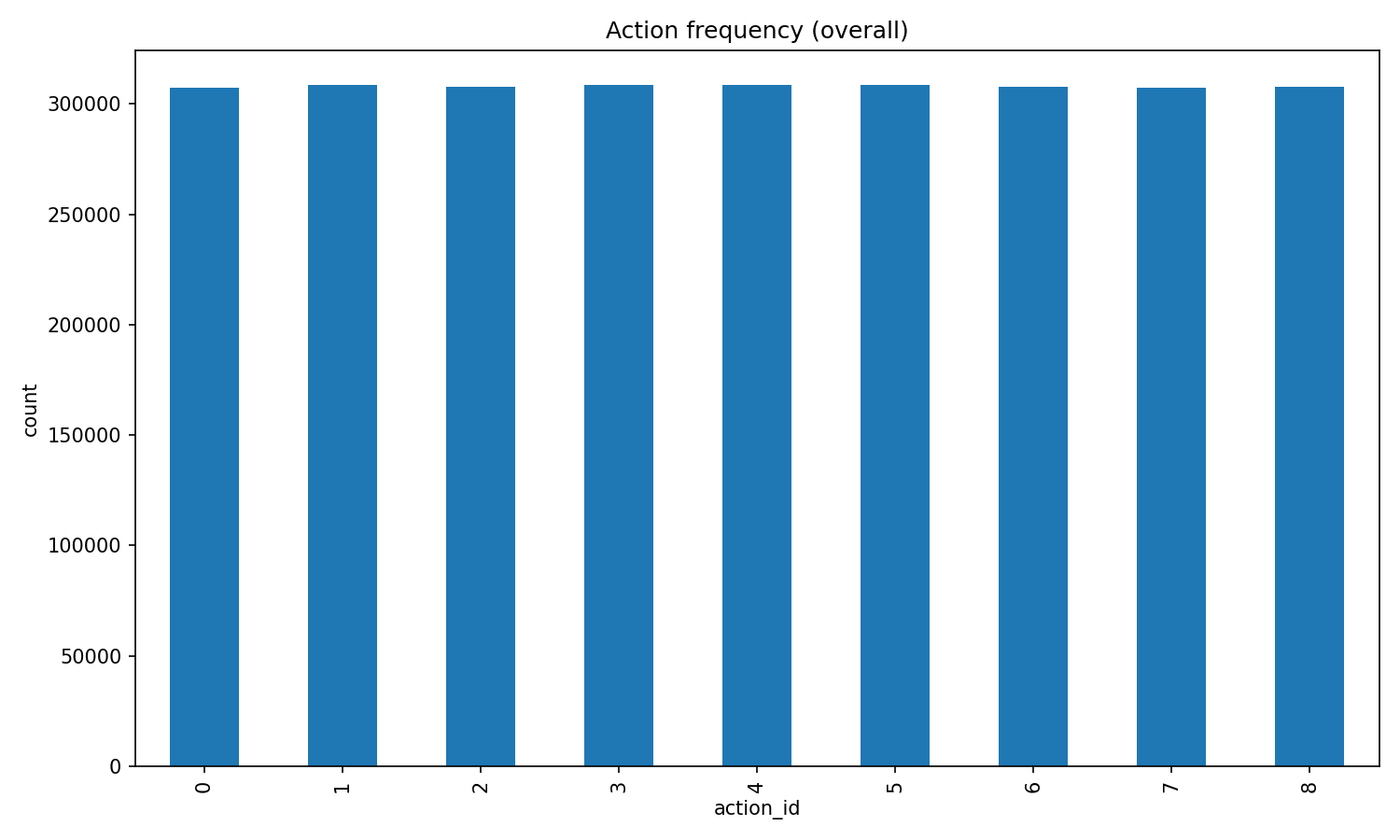}\hfill
  \includegraphics[width=0.32\linewidth]{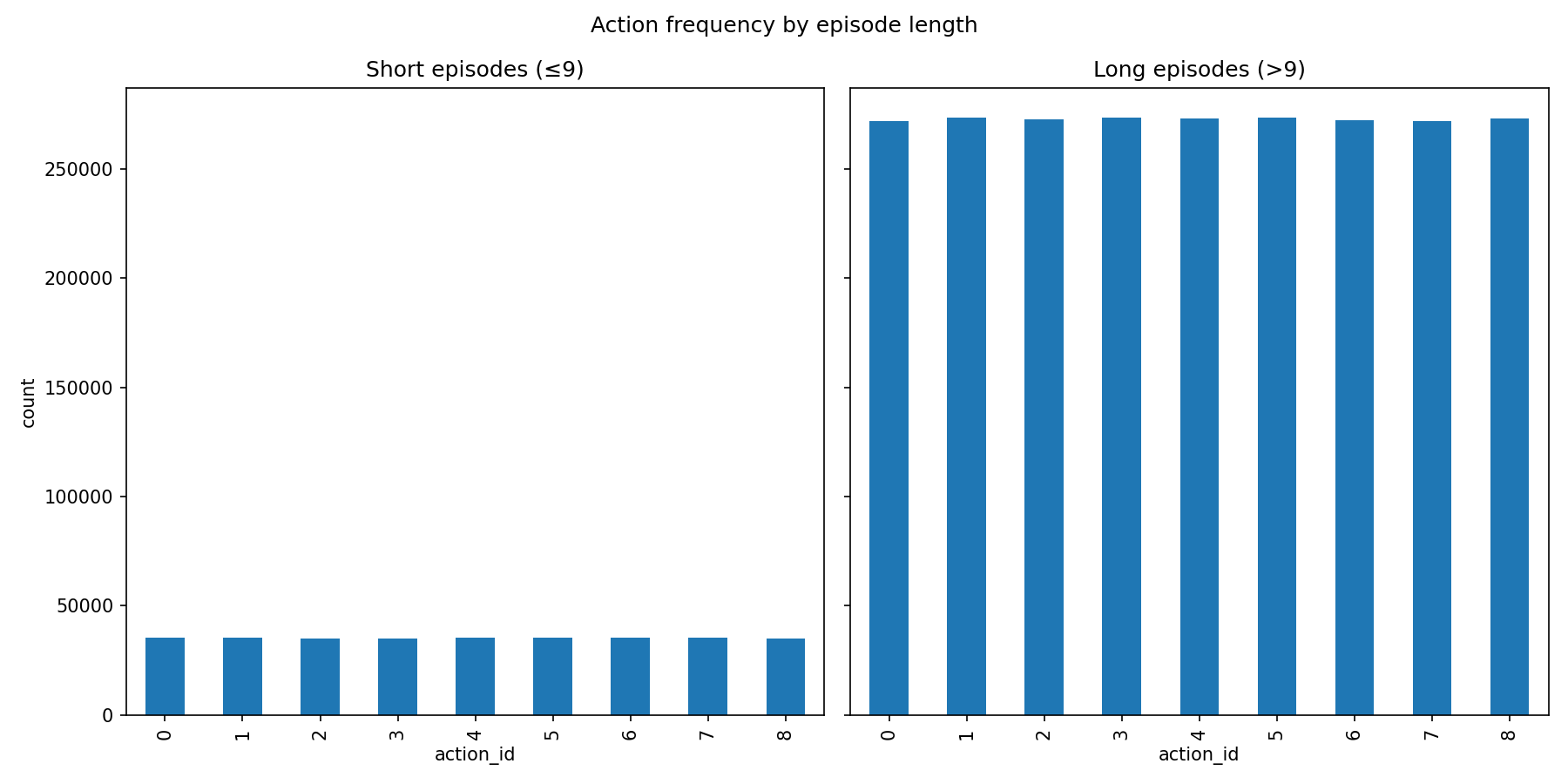}\hfill
  \includegraphics[width=0.32\linewidth]{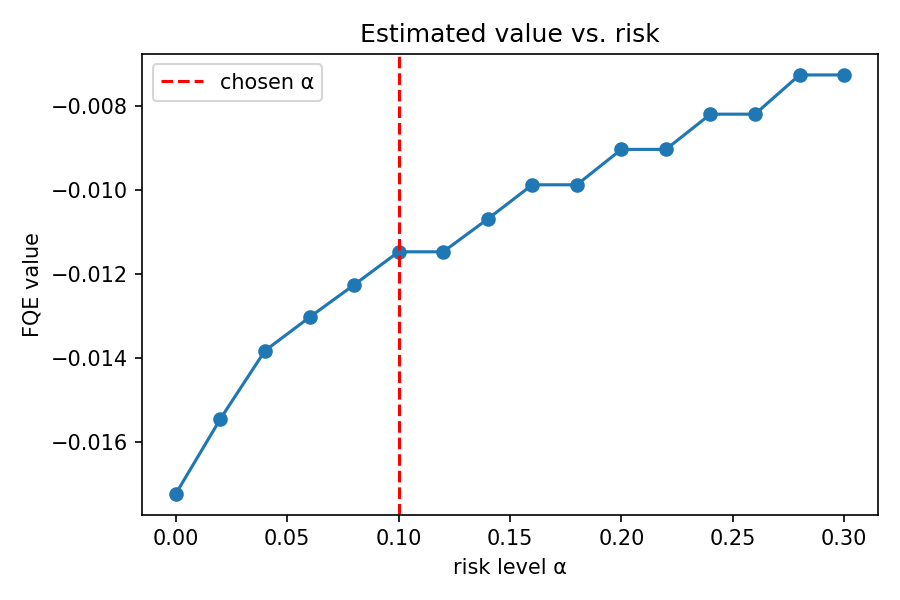}
  \vspace{-4pt}
  \caption{Supplementary plots. Left: action distribution overall. Middle: action histogram by episode length. Right: value versus $\alpha$ sensitivity.}
  \label{fig:supp_actions}
\end{figure}

\begin{figure}[t]
  \centering
  \includegraphics[width=0.32\linewidth]{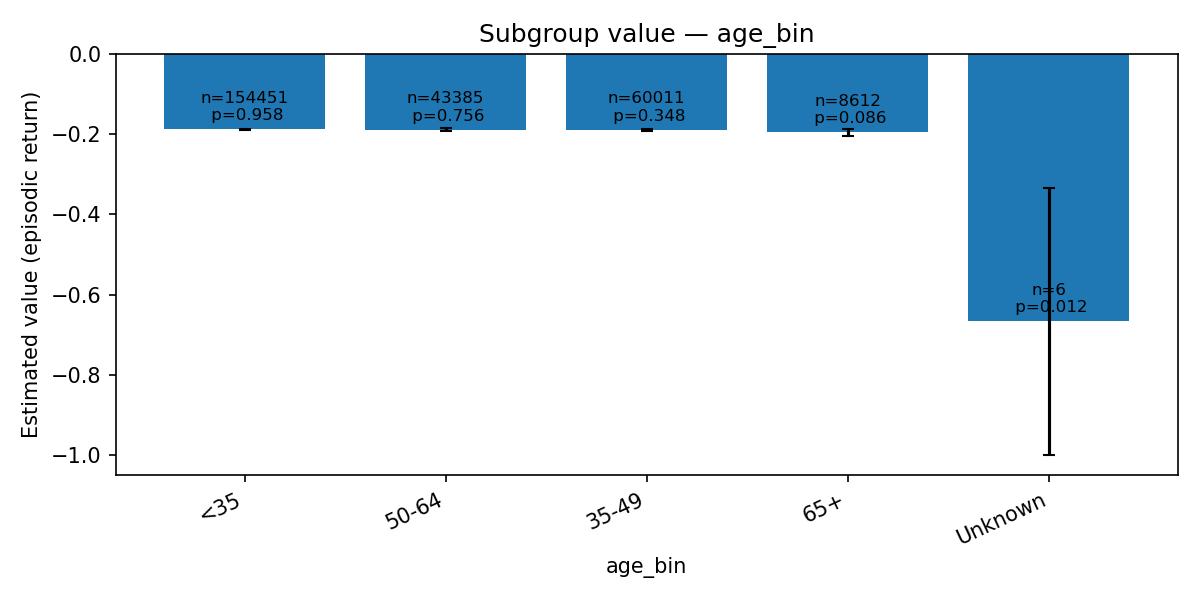}
  \includegraphics[width=0.32\linewidth]{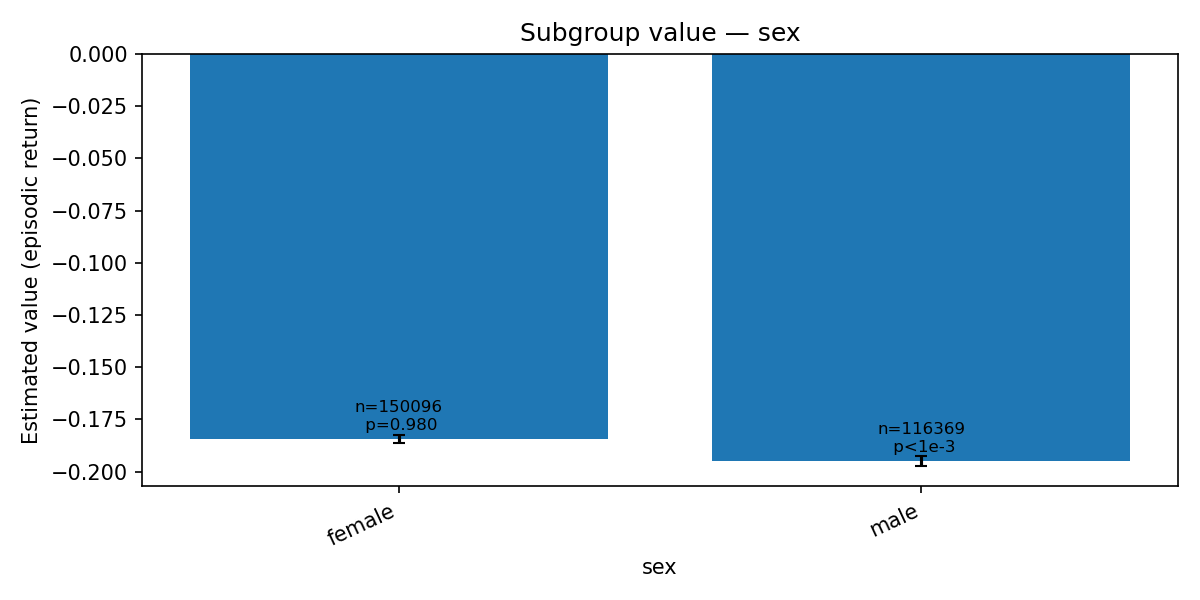}
  \includegraphics[width=0.32\linewidth]{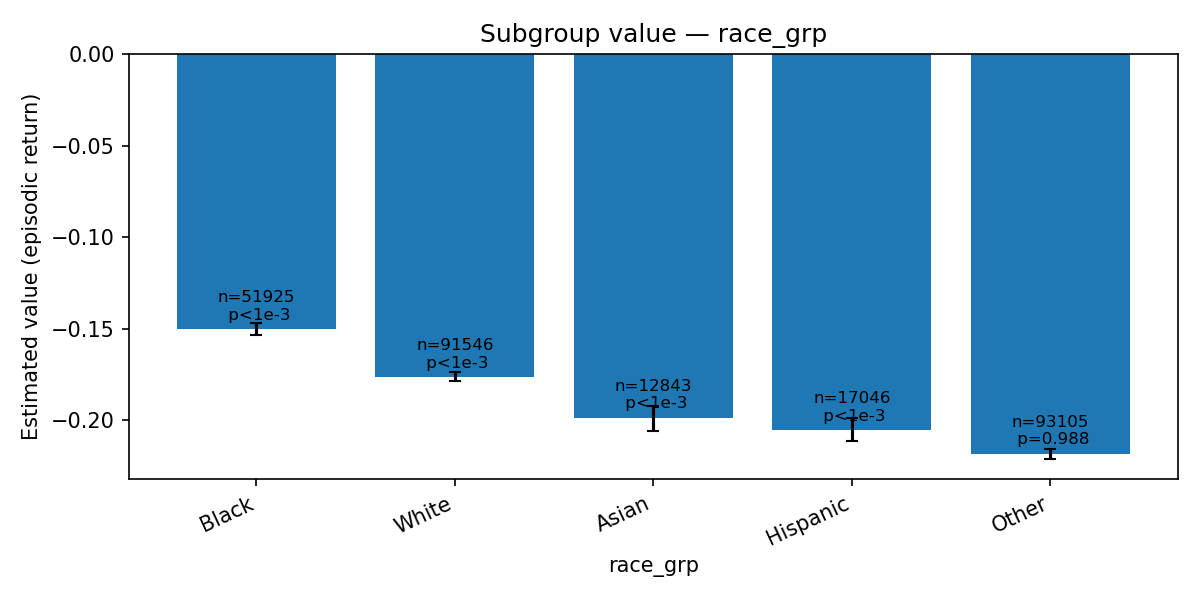}
  \vspace{-4pt}
  \caption{Supplementary FQE value summaries by subgroup (age, sex, race).}
  \label{fig:supp_fqe}
\end{figure}

\section*{Acknowledgements}
We thank the Waymark operations and engineering teams for data engineering support and domain expertise.

\bibliographystyle{unsrtnat}
\bibliography{refs}

\begin{thebibliography}{21}
\providecommand{\natexlab}[1]{#1}
\providecommand{\url}[1]{\texttt{#1}}
\expandafter\ifx\csname urlstyle\endcsname\relax
  \providecommand{\doi}[1]{doi: #1}\else
  \providecommand{\doi}{doi: \begingroup \urlstyle{rm}\Url}\fi

\bibitem[Levine et~al.(2020)Levine, Kumar, Tucker, and Fu]{levine2020offline}
Sergey Levine, Aviral Kumar, George Tucker, and Justin Fu.
\newblock Offline reinforcement learning: Tutorial, review, and perspectives on
  open problems.
\newblock \emph{arXiv:2005.01643}, 2020.

\bibitem[Kidambi et~al.(2020)Kidambi, Rajeswaran, Netrapalli, and
  Kakade]{kidambi2020morel}
Rahul Kidambi, Aravind Rajeswaran, Praneeth Netrapalli, and Sham Kakade.
\newblock Morel: Model-based offline reinforcement learning.
\newblock In \emph{NeurIPS}, 2020.

\bibitem[Vovk et~al.(2005)Vovk, Gammerman, and Shafer]{vovk2005algorithmic}
Vladimir Vovk, Alex Gammerman, and Glenn Shafer.
\newblock \emph{Algorithmic Learning in a Random World}.
\newblock Springer, 2005.

\bibitem[Angelopoulos and Bates(2023)]{angelopoulos2023conformal}
Anastasios~N Angelopoulos and Stephen Bates.
\newblock Conformal prediction: A gentle introduction.
\newblock \emph{Foundations and Trends in Machine Learning}, 16\penalty0
  (4):\penalty0 494--591, 2023.

\bibitem[{Centers for Medicare and Medicaid Services}(2022)]{cmsframework2022}
{Centers for Medicare and Medicaid Services}.
\newblock Cms framework for health equity 2022--2032.
\newblock
  \url{https://www.cms.gov/files/document/cms-framework-health-equity-2022.pdf},
  2022.

\bibitem[Fisch et~al.(2022)]{fisch2022calibrated}
Adam Fisch et~al.
\newblock Calibrated selective classification.
\newblock \emph{NeurIPS}, 2022.

\bibitem[Romano et~al.(2019)Romano, Patterson, and
  Candes]{romano2019conformalized}
Yaniv Romano, Evan Patterson, and Emmanuel Candes.
\newblock Conformalized quantile regression.
\newblock In \emph{NeurIPS}, 2019.

\bibitem[Kumar et~al.(2020)Kumar, Zhou, Tucker, and
  Levine]{kumar2020conservative}
Aviral Kumar, A~Zhou, George Tucker, and Sergey Levine.
\newblock Conservative q-learning for offline reinforcement learning.
\newblock In \emph{NeurIPS}, 2020.

\bibitem[Kostrikov et~al.(2021)Kostrikov, Nair, and
  Levine]{kostrikov2021offline}
Ilya Kostrikov, Ashvin Nair, and Sergey Levine.
\newblock Offline reinforcement learning with implicit q-learning.
\newblock In \emph{ICLR}, 2021.

\bibitem[{National Committee for Quality Assurance}(2024)]{ncqahedis2024}
{National Committee for Quality Assurance}.
\newblock Ncqa hedis health equity summary measures.
\newblock \url{https://www.ncqa.org/hedis/measures/}, 2024.

\bibitem[Fu et~al.(2021)Fu, Kumar, Nachum, Tucker, and Levine]{fu2021d4rl}
Justin Fu, Aviral Kumar, Ofir Nachum, George Tucker, and Sergey Levine.
\newblock D4rl: Datasets for deep data-driven reinforcement learning.
\newblock \emph{arXiv:2004.07219}, 2021.

\bibitem[Le et~al.(2019)Le, Voloshin, and Yue]{le2019batch}
Hoang~M Le, Cameron Voloshin, and Yisong Yue.
\newblock Batch policy learning under constraints.
\newblock In \emph{ICML}, 2019.

\bibitem[Jiang and Li(2016)]{jiang2016doubly}
Nan Jiang and Lihong Li.
\newblock Doubly robust off-policy value evaluation for reinforcement learning.
\newblock In \emph{ICML}, 2016.

\bibitem[Thomas et~al.(2015)Thomas, Theocharous, and
  Ghavamzadeh]{thomas2015high}
Philip Thomas, Georgios Theocharous, and Mohammad Ghavamzadeh.
\newblock High-confidence off-policy evaluation.
\newblock In \emph{AAAI}, 2015.

\bibitem[Uehara et~al.(2020)Uehara, Jiang, et~al.]{uehara2020minimax}
Masatoshi Uehara, Nan Jiang, et~al.
\newblock Minimax weight and q-function learning for off-policy evaluation.
\newblock In \emph{ICML}, 2020.

\bibitem[Shafer and Vovk(2008)]{shafer2008tutorial}
Glenn Shafer and Vladimir Vovk.
\newblock A tutorial on conformal prediction.
\newblock \emph{Journal of Machine Learning Research}, 9:\penalty0 371--421,
  2008.

\bibitem[Barocas and Selbst(2017)]{barocas2017problem}
Solon Barocas and Andrew~D Selbst.
\newblock The problem with bias: Allocative and representational harms in
  machine learning.
\newblock \emph{Big Data}, 2017.

\bibitem[Rajkomar et~al.(2018)]{rajkomar2018ensuring}
Alvin Rajkomar et~al.
\newblock Ensuring fairness in machine learning to advance health equity.
\newblock \emph{Annals of Internal Medicine}, 2018.

\bibitem[Jabbari et~al.(2017)Jabbari, Joseph, Kearns, Morgenstern, and
  Roth]{jabbari2017fairness}
Shahin Jabbari, Matthew Joseph, Michael Kearns, Jamie Morgenstern, and Aaron
  Roth.
\newblock Fairness in reinforcement learning.
\newblock In \emph{ICML Workshop on Fairness, Accountability, and
  Transparency}, 2017.

\bibitem[Heidari et~al.(2019)Heidari, Loi, Gummadi, and
  Krause]{heidari2019long}
Hoda Heidari, Michele Loi, Krishna Gummadi, and Andreas Krause.
\newblock On the long-term impact of algorithmic decision policies: Effort
  unfairness and status quo bias.
\newblock In \emph{NeurIPS}, 2019.

\bibitem[Obermeyer et~al.(2019)Obermeyer, Powers, Vogeli, and
  Mullainathan]{obermeyer2019dissecting}
Ziad Obermeyer, Brian Powers, Christine Vogeli, and Sendhil Mullainathan.
\newblock Dissecting racial bias in an algorithm used to manage population
  health.
\newblock \emph{Science}, 366\penalty0 (6464):\penalty0 447--453, 2019.

\end{thebibliography}

\end{document}